\documentclass[10pt,twocolumn,letterpaper]{article}

\usepackage{cvpr}
\usepackage{times}
\usepackage{epsfig}
\usepackage{graphicx}
\usepackage{amsmath}
\usepackage{amssymb}

\usepackage{algorithmic}
\usepackage{algorithm}

\usepackage[pagebackref=true,breaklinks=true,citecolor=blue,linkcolor=blue,letterpaper=true,colorlinks,bookmarks=false]{hyperref}

 \cvprfinalcopy 

\def\cvprPaperID{2359} 

\ifcvprfinal\fi
\begin{document}
%
\title{Learning Compact Target-Oriented Feature Representations for Visual Tracking}

\author{Chenglong Li$^{1,2,3}$, Yan Huang$^{3}$, Liang Wang$^{3}$, Jin Tang$^{1}$ and Liang Lin$^{4}$\\
$^1$Key Laboratory of Intelligent Computing and Signal Processing of Ministry of Education,\\
School of Computer Science and Technology, Anhui University\\
$^2$Institute of Physical Science and Information Technology, Anhui University\\
$^3$Center for Research on Intelligent Perception and Computing, NLPR, CASIA\\
School of Data and Computer Science, Sun Yat-Sen University
\\
{\tt\small lcl1314@foxmail.com, \{yhuang,wangliang\}@nlpr.ia.ac.cn, tangjin@ahu.edu.cn, linliang@ieee.org}
}

\maketitle
\begin{abstract}
Many state-of-the-art trackers usually resort to the pretrained convolutional neural network (CNN) model for correlation filtering, in which deep features could usually be
redundant, noisy and less discriminative for some certain instances, and the tracking performance might thus be affected. To handle this problem, we propose a novel approach, which takes both advantages of good generalization of generative models and excellent discrimination
of discriminative models, for visual tracking. In particular,
we learn compact, discriminative and target-oriented feature representations using the Laplacian coding algorithm that exploits the dependence among the input local features in a discriminative correlation filter framework. The feature representations
and the correlation filter are jointly learnt to enhance to each
other via a fast solver which only has very slight computational burden on the tracking speed. Extensive experiments on
three benchmark datasets demonstrate that this proposed framework
clearly outperforms baseline trackers with a modest impact
on the frame rate, and performs comparably against the state-of-the-art methods.
\end{abstract}

\section{Introduction}
\label{sec::introduction}
Given a video sequence, the task of visual tracking is
to locate an object instance whose state is specified at the
first frame. In general, tracking models can be grouped into
three categories, i.e., generative, discriminative and hybrid. Generative models aim to locate an image region
that is most similar to the target appearance, and possess a good
generalization when only a limited number of training samples are available~\cite{Ng01nips}. While discriminative ones are to
train binary classifiers to distinguish the target from background, and would achieve excellent performance if the size
of the training set is sufficiently large~\cite{Lasserrre06cvpr}. Hybrid models are
usually inherited from both advantages of generative and
discriminative models~\cite{SCM12cvpr,Sui15iccv}.

\begin{figure}[t]
  \centering
  \includegraphics[width=0.9\columnwidth]{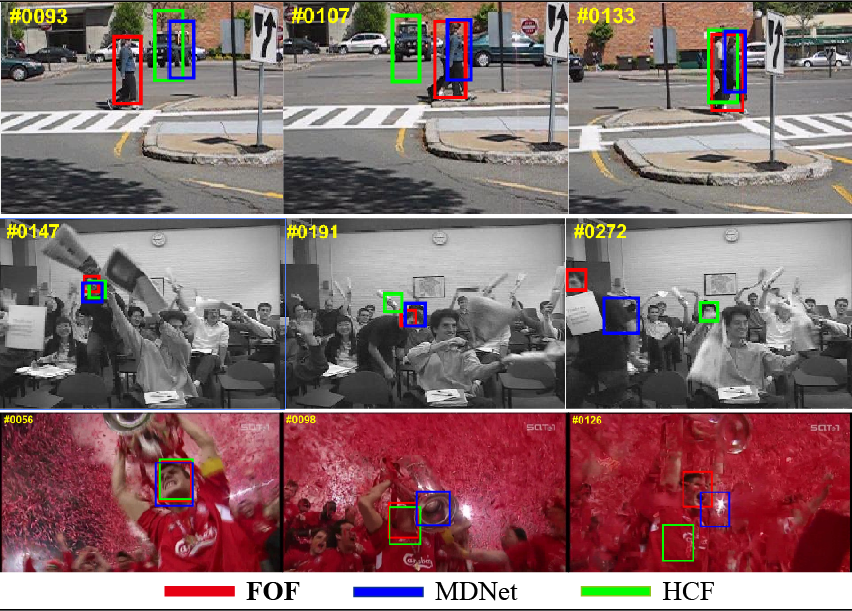} \\
  \caption{A comparison of our approach FOF with the baseline
HCF~\cite{Ma15iccv} and the state-of-the-art MDNet~\cite{MDNet15cvpr} on three example
sequences. Our FOF tracker successfully tackles the challenges of
motion blur, low resolution, partial occlusion, appearance variation and background clutter. }\label{fig::comparison_results}
\end{figure}

Recent studies on tracking are dominated by deep convolutional neural network (CNN)~\cite{Ma15iccv,C-COT16eccv,MDNet15cvpr,ECO17cvpr,StrutSiam18eccv}, and some of them resort
to the pre-trained CNN model for correlation filtering. Correlation filter based tracking models exploit circular shifts to generate thousands of translated training samples, and put
training and detection into the Fourier domain according to
the circulant properties of the translated sample features, which can
reduce both storage and computation by several orders of
magnitude~\cite{CSK15pami}. Although achieving appealing results, most
of these correlation filter based methods directly use feature maps for training samples (e.g., CNN features), and
their performance in robustness might be affected by redundancy, noise and less
discriminative ability for some certain instances of deep features. For example, the output of the \emph{conv5-4} convolutional layer in the VGGNet-19~\cite{vgg15iclr} trained on ImageNet is widely used for visual tracking and has 512 dimensions, most of whose elements are zeros and scattered~\cite{ECO17cvpr}. In addition, the above feature may be sufficient to represent generic
targets, but its effectiveness in terms of tracking
is limited due to the fundamental inconsistency between classification and tracking problems, i.e., predicting object
class labels versus locating targets of arbitrary classes~\cite{MDNet15cvpr}.

The Bag-of-features (BoF) framework is widely used in various applications, such as image classification and retrieval~\cite{LSC13pami,Huang14pami}. It encodes each input feature on the constructed codebook into a
coding vector, which not only benefits from the representation ability of original feature, but also is more compact and
discriminative~\cite{SCM12cvpr,Wang15tip}. The usage of BoF in visual tracking is to reconstruct the input features on the positive and negative dictionaries, and then employ the reconstruction coefficients (i.e., coding features)~\cite{SCM12cvpr,Liu11cvpr,DSSM14tip,Zhang17tnnls} to define the likelihood scores of candidates in the Bayesian or particle filtering framework. For these methods, however, there are two major issues to be not addressed yet. First, the used features are pixel intensities which are too weak, significantly limiting the tracking performance. In addition, these methods usually
requires solving the $\ell_1$-minimization problem as many times
as the number of candidates. Considering the time nature of visual tracking, therefore, it is unacceptable to use very high dimensional deep features in this framework. Second, the sparse sampling strategy is difficult to balance the trade-off between tracking accuracy and computational burden.

This paper takes both advantages of discriminative and generative models into account for visual tracking, and handles all of above-mentioned problems of correlation filter and BoF based tracking algorithms .
In particular, we elaborately design a novel hybrid model to
enhance the discriminative capability of the correlation filter with the usage of good generalization of feature
coding to mitigate the redundancy and noise effects of deep features, and thus achieve clearly improved tracking performance with the considerable efficiency. Specifically, we encode input features as new representations in a robust way using
the Laplacian coding algorithm that exploits the dependence among the local features, and learn it together with
the correlation filter in a single unified optimization framework. In addition to the \emph{compact} and \emph{discriminative} properties inherited from original feature coding methods, our
learnt feature representations are \emph{target-oriented} due to the
proposed joint optimization scheme, and thus significantly
augment the discriminative capability of the correlation filter. Some tracking examples are presented in Fig.~\ref{fig::comparison_results} to show the effectiveness of the proposed approach against the baseline and state-of-the-art trackers.

To our best knowledge, it is probably the first work to improve the correlation filter framework from the perspective of BoF, and we think it would be a
potential direction to the visual tracking task. We summarize the major contributions of this work as follows. 

\begin{itemize}
\item We propose an effective approach to alleviate the effects of feature redundancy and noise in visual tracking. Extensive experiments show that the proposed
method clearly outperforms baseline trackers with a
modest impact on the frame rate, and performs comparably against the state-of-the-art trackers on three benchmark datasets. Source codes and experimental results
would be available online for reproducible research.

\item We present a novel correlation filter model to augment
the discriminative ability by learning a compact, discriminative and target-oriented feature representations. The proposed model jointly optimizes the correlation filter and
the feature representations in a unified framework.

\item We develop an efficient algorithm to solve the associated optimization problem, where each sub-problem is
convex and the convergence is guaranteed. Moreover, we analyze the computational complexity of the proposed algorithm in detail. Empirically,
our algorithm can converge within very few iterations
on real image data, and thus only has very slight computational burden on the tracking speed.

\end{itemize}

\section{Related Work}
\label{sec::related_work}
There are many kinds of trackers~\cite{Li16tip1,Li18eccv,Li18pami,Li19pr}, and here we only introduce some methods most relevant to ours in this section.

\subsection{Correlation Filter for Tracking}
The early work of correlation filters (CFs) for tracking is MOSSE~\cite{MOSSE10cvpr} which uses a set of training samples to train CFs in the frequency domain. Henriques~\emph{et al.}~\cite{CSK15pami} extend CFs with the kernel trick and multi-channel features. Based on these works, several notable improvements have been proposed. For example, scale-adaptive schemes are incorporated for handling {\bf scale variation}~\cite{DSST17pami,arv17iccvw}, and part-based models are proposed for addressing {\bf partial occlusion}~\cite{liu15cvpr,liu16cvpr}. To utilize the complementary benefits of different features, multiple {\bf features integration}~\cite{Ma15iccv,bertinetto2016staple,C-COT16eccv,ECO17cvpr} are fully investigated for improving tracking performance. To mitigate background effects, {\bf background context} is used to enhance the discriminative ability of correlation filters, i.e., suppressing background information in filter learning~\cite{ba17iccv,ca17cvpr}. In addition, various spatially regularized models are proposed to handle {\bf boundary effect}~\cite{danelljan2015learning,dcf-csr16cvpr} caused by periodic repetitions of circulant shifted samples.

\subsection{Feature Coding for Tracking}
Feature coding (FC) is a core component of the BoF framework~\cite{LSC13pami,Huang14pami}, and has been widely applied in different fields of computer vision, including the tracking task.
Inspired by the properties of the receptive fields of simple cells in visual cortex, FC uses the representation coefficients as features to describe the appearance of target
objects~\cite{Zhang13pr}. Liu et al.~\cite{Liu11cvpr} propose a local sparse appearance model to learn a target-based sparse coding histogram,
and then employ the meanshift algorithm to perform tracking. Zhong et al.~\cite{SCM12cvpr} propose a sparsity-based collaborative
model in the Bayesian filtering framework, in which a sparsity-based generative model is developed to construct histogram
features for target objects, and spatial information and occlusion handling are incorporated. A biologically inspired method is proposed by Zhang et al.~\cite{Zhang17tnnls}
to model the target appearance via a coding layer, and tracking
is carried out in a particle filter framework. With the same
tracking framework, other coding algorithms are employed
to design effective trackers, such as locality-constrained linear coding~\cite{Ding18iet} and sparse and local linear coding~\cite{Wang15tip}. Different from them, we jointly learn the feature code
and the correlation filter in a unified optimization framework so as to yield a more compact, discriminative and target-oriented feature representation.

\section{Dual Correlation Filter}
\label{sec::preliminaries}
In this section, we give a brief description of correlation filter (CF) and its dual form (DCF), which is preliminary for our algorithm. The key idea of CF is that thousands of negative samples have the circulant structure whose computation can be transformed in the Fourier domain at a high speed~\cite{CSK15pami}. 

Given an image patch ${\bf x}$ with the size of $M\times N$, CF trackers use all circular shifts, denoting as ${\bf x}_{m,n}$, to train a correlation filter ${\bf w}$, where $(m,n)\in\{0,1,...,M-1\}\times\{0,1,...,N-1\}$. The labels ${\bf y}_{m,n}$ of these shifted samples are generated by a Gaussian function, and the goal is to find the optimal weights ${\bf w}$ in the following program:
\begin{equation}\label{eq::cf}
\begin{aligned}
&\arg\min_{\bf w}~\sum_{m,n}\lVert{\bf c}_{m,n}\odot{\bf x}_{m,n}\odot{\bf w}_{m,n}-{\bf y}_{m,n}\rVert^2_2+\lambda\lVert{\bf w}\rVert^2_2,
\end{aligned}
\end{equation}
where $\lVert\cdot\rVert_2^2$ denotes the $\ell_2$-norm of a vector, and $\lambda$ is a regularization parameter. $\odot$ indicates the element-wise product, and ${\bf c}$ is a cosine window used to suppress boundary effects of periodic shift samples~\cite{CSK15pami}. To simplify computation for multiple channel features, the dual form of~\eqref{eq::cf}, called dual correlation filter (DCF), can equivalently be expressed as follows:
\begin{equation}\label{eq::dcf}
\begin{aligned}
&\arg\min_{\bf u}~\frac{1}{4\lambda}{\bf u}^\top S({\bf x}_{\bf c})S({\bf x}_{\bf c})^\top{\bf u}+\frac{1}{4}{\bf u}^\top{\bf u}-{\bf u}^\top{\bf y},
\end{aligned}
\end{equation}
where ${\bf u}$ is the dual variable of ${\bf w}$, $S({\bf x})$ denotes a circulant matrix whose base vector is ${\bf x}$, and ${\bf c}\odot{\bf x}$ is denoted as ${\bf x}_{\bf c}$ for simplicity. Through using the fast Fourier transformation (FFT) to diagonalize the circulant matrix, the solution of~\eqref{eq::dcf} is as:
\begin{equation}\label{eq::solution-dcf}
\begin{aligned}
&\hat{\bf u}=\frac{\hat{\bf y}}{\frac{1}{2\lambda}\hat{\bf x}^*_{\bf c}\odot\hat{\bf x}_{\bf c}+\frac{1}{2}},
\end{aligned}
\end{equation}
where $\hat{\bf u}$ denotes the discrete Fourier transformation (DFT) of ${\bf u}$, i.e., $\hat{\bf u}=\mathcal{F}({\bf u})$, and $\mathcal{F}(\cdot)$ represents the Fourier transformation. ${\bf x}^*$ is the complex-conjugate of ${\bf x}$. ${\bf u}$ can be obtained via ${\bf u}=\mathcal{F}^{-1}(\hat{\bf u})$, where $\mathcal{F}(\cdot)$ indicates the inverse Fourier transformation. If ${\bf x}$ has $D$-dimensional channels, by simply summing over them in the Fourier domain~\cite{CSK15pami}, the solution of~\eqref{eq::dcf} can be written as:
\begin{equation}\label{eq::solution-dcf1}
\begin{aligned}
&\hat{\bf u}=\frac{\hat{\bf y}}{\frac{1}{2\lambda}\sum_d\hat{\bf x}^{d*}_{\bf c}\odot\hat{\bf x}^d_{\bf c}+\frac{1}{2}},
\end{aligned}
\end{equation}
where $d\in\{1,2,...,D\}$ denotes the channel index.

\section{Filter Optimization Driven Feature Coding}
\label{sec::joint-learning}
In this section, we will describe the proposed model in detail,
and present the associated optimization algorithm.

\subsection{Model Formulation}
As discussed above, we aim at using feature coding-based object representation to enhance the discriminative
capacity of dual correlation filter (DCF) while utilizing the
filter optimization to guide feature learning.

\begin{figure}[t]
  \centering
  \includegraphics[width=0.7\columnwidth]{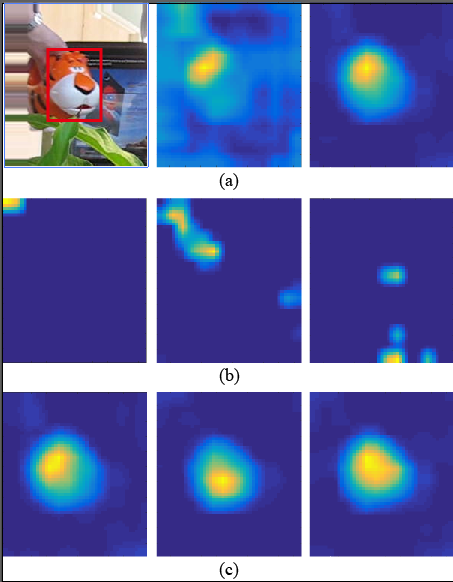} \\
  \caption{Illustration of feature maps from the output (denoted by
${\bf X}$) of the \emph{conv5-4} layer in VGGNet-19~\cite{vgg15iclr} trained on ImageNet and
the learnt feature representation (denoted by ${\bf Z}$). (a) From left to
right: search window of a target object (indicated by red bounding
box), average feature map of ${\bf X}$, and average feature map of ${\bf Z}$,
where the input image patch is from the \emph{tiger2} sequence in the
OTB100 dataset~\cite{RGBbenchmark15pami}. (b) Three feature maps randomly selected
from ${\bf X}$. (c) Three feature maps randomly selected from ${\bf Z}$. One
can see that the learnt features are more compact, discriminative
and target-oriented than those directly extracted from VGGNet-19.}\label{fig::illustration_feature_maps}
\end{figure}

A set of codewords needs to be generated first to
compose a codebook ${\bf B}\in\mathcal{R}^{D\times k}$, where $D$ is the feature
dimension, and $k$ is the number of codebook elements. We will discuss the details of cookbook later. We encode the input target patch feature ${\bf x}$ on ${\bf B}$, and integrate the encoded feature representation into the dual correlation filter
(DCF). To this end, we rearrange the multi-channel feature
vector ${\bf x}\in\mathcal{R}^{MN\times D}$ as a matrix ${\bf X}\in\mathcal{R}^{D\times MN}$, and represent it using the coefficient ${\bf Z}\in\mathcal{R}^{k\times MN}$ on ${\bf B}$: ${\bf X}={\bf B}{\bf Z}$.
Notice that $k$ can be viewed as the channel number of ${\bf Z}$,
and we use $k = 1$ to consider the single channel case for
the convenience of our formulation. When $k > 1$, the
solution of ${\bf u}$ can be obtained by simply summing over all
channels of ${\bf Z}$ in the Fourier domain, as discussed in Section~\ref{sec::preliminaries}. Instead of using original feature ${\bf x}$, we employ the
more compact and discriminative feature coding representation ${\bf Z}$~\cite{Huang14pami,Jegou12pami} to learn DCF in~\eqref{eq::dcf}, and formulate it as follows:
\begin{equation}\label{eq::combine}
\begin{aligned}
& \min_{{\bf Z},{\bf u}} \frac{1}{4\lambda}{\bf u}^\top S({\bf Z}_{\bf c})S({\bf Z}_{\bf c})^\top{\bf u}+\frac{1}{4}{\bf u}^\top{\bf u}
-{\bf u}^\top{\bf y},\\
&+\Phi({\bf Z})~s.t.~{\bf X}={\bf B}{\bf Z},
\end{aligned}
\end{equation}
where $\Phi({\bf Z})$ represents prior constraints on ${\bf Z}$. As seen
from~\eqref{eq::combine}, in addition to the compact and discriminative
properties inherited from the feature coding algorithm, the
learnt feature ${\bf Z}$ is also target-oriented due to the joint learning scheme of ${\bf Z}$ and ${\bf u}$. In addition, if we use the pre-trained
CNN features, the model in~\eqref{eq::combine} can significantly reduce the
feature dimension to remove redundancy and noise in learning the filter. For example, the output of the \emph{conv5-4} convolutional layer widely used in visual tracking has 512 dimensions while the dimension $k$ of ${\bf Z}$ is set to 10 in this
work. Although the dimension of the learnt feature (i.e.,
${\bf Z}$) is much smaller than the \emph{conv5-4} convolutional feature
(i.e., ${\bf X}$), ${\bf Z}$ is more discriminative and target-oriented than
${\bf X}$, as shown in Fig.~\ref{fig::illustration_feature_maps}. Moreover, we can observe that different feature maps from ${\bf Z}$ mainly focus on different parts of the target object while suppressing background parts, which can be explained by the fact that the proposed joint
learning algorithm strengthens the discriminative power of feature representations. Therefore, the learnt features make the filter more robust to
various challenges, such as appearance change and background clutter.

Various prior constraints could be explored to regularize ${\bf Z}$
for better stability and the quality of the feature coding, such as Frobenius norm, low rank and sparse, and we adopt simple yet effective one for computational efficiency. Note that similar features
may be encoded as totally different sparse codes, and such instability easily harms the
robustness of the feature coding~\cite{LSC13pami} and thus might affect the tracking performance. Therefore, we put the Laplacian constraint on ${\bf Z}$ that preserves
the locality and similarity information among local features to alleviate the instability of feature coding, and the final
joint learning model is as follows:
\begin{equation}\label{eq::objective}
\begin{aligned}
& \min_{{\bf Z},{\bf u}} \frac{1}{4\lambda}{\bf u}^\top S({\bf Z}_{\bf c})S({\bf Z}_{\bf c})^\top{\bf u}+\frac{1}{4}{\bf u}^\top{\bf u}
-{\bf u}^\top{\bf y}\\
&+\gamma~tr({\bf ZLZ}^\top),~s.t.~{\bf X}={\bf B}{\bf Z},
\end{aligned}
\end{equation}
where $tr(\cdot)$ indicates the matrix trace, and $\gamma$ is the balanced parameter. ${\bf L}={\bf F}-{\bf G}$ is the Laplacian matrix, where ${\bf F}$ is the degree matrix whose diagonal element $F_i=\sum_jG_{ij}$, and ${\bf G}$ is a binary matrix indicating the relationship between any two coding features with $G_{ij}=1$ if ${\bf z}_i$ is among the $r$ nearest neighbors of ${\bf z}_j$ otherwise  $G_{ij}=0$.

\subsection{Optimization Algorithm}
The model in~\eqref{eq::objective} is not joint convex on ${\bf Z}$, ${\bf E}$ and ${\bf u}$, but it is convex with respect to each of them when others are fixed. The ADMM (alternating direction method of multipliers) algorithm~\cite{ADMM11} has shown to be an efficient and effective solver of such problems. To apply ADMM for the above problem, we need to make the objective function separable. Therefore, we introduce an auxiliary variable ${\bf p}$ to replace ${\bf Z}_{\bf c}$ in~\eqref{eq::objective}:
\begin{equation}\label{eq::separable}
\begin{aligned}
& \min_{{\bf Z},{\bf u}} \frac{1}{4\lambda}{\bf u}^\top S({\bf p})S({\bf p})^\top{\bf u}+\frac{1}{4}{\bf u}^\top{\bf u}
-{\bf u}^\top{\bf y}\\
&+\gamma~tr({\bf ZLZ}^\top),~s.t.~{\bf X}={\bf B}{\bf Z},{\bf p}={\bf Z}_{\bf c}.
\end{aligned}
\end{equation}

The augmented Lagrangian function is:
\begin{equation}\label{eq::lagrangian}
\begin{aligned}
& \mathcal{L}_{\{{\bf Z},{\bf u},{\bf p}\}} =\frac{1}{4\lambda}{\bf u}^\top S({\bf p})S({\bf p})^\top{\bf u}+\frac{1}{4}{\bf u}^\top{\bf u}
-{\bf u}^\top{\bf y}\\
&+\frac{\mu}{2}\lVert{\bf X}-{\bf B}{\bf Z}\rVert_F^2+\gamma~tr({\bf ZLZ}^\top)+\langle{\bf Y}_1,{\bf X}-{\bf B}{\bf Z}\rangle\\
&+\frac{\mu}{2}\lVert{\bf X}-{\bf B}{\bf Z}\rVert_2^2+\langle{\bf y}_2,{\bf p}-{\bf Z}_{\bf c}\rangle+\frac{\mu}{2}\lVert{\bf p}-{\bf Z}_{\bf c}\rVert_2^2,
\end{aligned}
\end{equation}
where ${\bf Y}_1$ and ${\bf y}_2$ are the Lagrange multipliers, and $\mu$ is the Lagrange parameter. The augmented Lagrangian function~\eqref{eq::lagrangian} can be iteratively minimized by ADMM which sequentially solves the following sub-problems at each iteration:
\begin{equation}\label{eq::Z}
\begin{aligned}
&\min_{{\bf Z}}~\frac{\mu}{2}\lVert{\bf X}-{\bf B}{\bf Z}+\frac{{\bf Y}_1}{\mu}\rVert_F^2+\frac{\mu}{2}\lVert{\bf p}-{\bf Z}_{\bf c}+\frac{{\bf y}_2}{\mu}\rVert_2^2\\
&+\gamma~tr({\bf ZLZ}^\top),
\end{aligned}
\end{equation}
\begin{equation}\label{eq::P}
\begin{aligned}
&\min_{{\bf p}}\frac{1}{4\lambda}{\bf u}^\top S({\bf p})S({\bf p})^\top{\bf u}+\frac{\mu}{2}\lVert{\bf p}-{\bf Z}_{\bf c}+\frac{{\bf y}_2}{\mu}\rVert_2^2,
\end{aligned}
\end{equation}
\begin{equation}\label{eq::u}
\begin{aligned}
&\min_{{\bf u}}\frac{1}{4\lambda}{\bf u}^\top S({\bf p})S({\bf p})^\top{\bf u}+\frac{1}{4}{\bf u}^\top{\bf u}
-{\bf u}^\top{\bf y}.
\end{aligned}
\end{equation}

{\flushleft \bf Efficient solutions}.
The problem in~\eqref{eq::Z} is a convex problem,
but does not have a closed-form solution. In this work, we
solve it efficiently using the Nesterov's Accelerated Gradient descent (NAG) algorithm~\cite{NAG16arXiv}. ${\bf p}$ in~\eqref{eq::P} and ${\bf u}$ in~\eqref{eq::u}
can be calculated fast in the Fourier domain~\cite{CSK15pami}. With simple algebra, the solutions of the above sub-problems are as follows:
\begin{equation}\label{eq::Z-solution}
\begin{aligned}
&{\bf Z}=\mathcal{N}(f({\bf Z})),
\end{aligned}
\end{equation}
\begin{equation}\label{eq::P-solution}
\begin{aligned}
&\hat{\bf p}=\frac{\mu\hat{\bf Z}_{\bf c}-\hat{\bf y}_2}{\frac{1}{2\lambda}\hat{\bf u}^*\odot\hat{\bf u}+\mu},
\end{aligned}
\end{equation}
\begin{equation}\label{eq::u-solution}
\begin{aligned}
&\hat{\bf u}=\frac{\hat{\bf y}}{\frac{1}{2\lambda}\hat{\bf p}^*\odot\hat{\bf p}+\frac{1}{2}},
\end{aligned}
\end{equation}
where $\mathcal{N}(\cdot)$ indicates the operator of the NAG, and $f({\bf Z})=\frac{\mu}{2}\lVert{\bf X}-{\bf B}{\bf Z}+\frac{{\bf Y}_1}{\mu}\rVert_F^2+\frac{\mu}{2}\lVert{\bf p}-{\bf Z}_{\bf c}+\frac{{\bf y}_2}{\mu}\rVert_2^2+\gamma~tr({\bf ZLZ}^\top)$.

The Lagrange multipliers and parameters are updated by a standard scheme~\cite{ADMM11}:
\begin{equation}\label{eq::multipliers}
\begin{aligned}
&{\bf Y}_1={\bf Y}_1+\mu({\bf X}-{\bf B}{\bf Z});\\
&{\bf y}_2={\bf y}_2+\mu({\bf p}-{\bf Z}_{\bf c});\\
&\mu=\min(\mu_m,\rho\mu),
\end{aligned}
\end{equation}
where $\mu_m$ denotes the maximum value of $\mu$ and $\rho$ is the scale parameter. 

\subsection{Discussion}
{\flushleft \bf Codebook construction}.
There are many methods for
codebook construction, e.g., $k$-means clustering and dictionary learning. $k$-means clustering is to initialize some
cluster centers randomly and then perform clustering using
the $k$-means algorithm to obtain the final clusters as dictionary elements. However, the quality of dictionary is greatly
affected by initial centers and the results are undeterministic as initial centers are randomly generated. Therefore,
we use the dictionary learning algorithm proposed in~\cite{Jenatton10icml}
to construct the codebook.
Considering the time-sensitive nature of visual tracking,
we construct the codebook in the first frame, and do not
update it in subsequent frames. On one hand, the target
image region in the first frame consists of the most representative patterns of the target object. On the other hand,
although new pattern of the target object appears, the pattern across different frames would encode similar features on
the fixed dictionary, and thus the tracking performance is
not affected much.

{\flushleft \bf Complexity analysis}.
Since ${\bf L}$ and ${\bf Z}$ are sparse matrices, and the computational cost of solving ${\bf Z}$ is
$\mathcal{O}(k^3N_JDMN)$, where $N_J$ is the maximum number of iterations of the NAG. The complexity of solving $\hat{\bf p}$ and $\hat{\bf s}$
is $\mathcal{O}(MN)$. Taking the FFT and inverse FFT into account, the complexity of solving $\hat{\bf p}$ and $\hat{\bf s}$ is
$\mathcal{O}(kMN log(MN))$. Hence, the overall cost of our algorithm is $\mathcal{O}(MN(k^3N_JD + kN_I log(MN)))$, where $N_I$ is
the maximum number of iterations of the ADMM. While
the complexity of DCF is $\mathcal{O}(MN log(MN))$. Since $k$, $N_I$
and $N_J$ are very small and $D$ is very smaller than $MN$, the
complexity of our algorithm is comparable with DCF. Note
that ${\bf B}^\top{\bf B}$ and ${\bf B}^\top{\bf X}$ can be precomputed, and the computational time is thus further reduced.

{\flushleft \bf Convergence}.
Note that each subproblem in~\eqref{eq::lagrangian} is convex,
and thus we can guarantee that the limit point by our algorithm satisfies the Nash equilibrium conditions~\cite{Xu13siam}. In
addition, we empirically find that the proposed optimization algorithm can converge within 2 iterations in ADMM
and 3 iterations in NAG on most of sequences, and thus set
$N_I$ to 2 in ADMM and $N_J$ to 3 in NAG for efficiency.

\begin{algorithm}[t]
\caption{Our Proposed Object Tracking Algorithm}\label{alg::tracking}
\begin{algorithmic}[1]
\REQUIRE
Input video sequence, target bounding box $bb_0$.
\ENSURE
Estimated target bounding box $bb^*_t$.
\STATE {\tt // Initialization}
\STATE Construct codebook ${\bf B}^l$ for the $l$-th layer and Guassian shape label vector ${\bf y}$;
\REPEAT
\STATE {\tt // Feature extraction}
\STATE Extract hierarchical convolutional features ${\bf X}_t^l$ and HOG feature ${\bf H}_t$ according to $bb_{t-1}$, and compute
Laplacian matrix ${\bf L}^l_t$ using ${\bf X}^l_t$;
\STATE {\tt // Target localization}
\STATE Solve~\eqref{eq::objective} with ${\bf B}^l$, ${\bf y}$, ${\bf X}_t^l$ and ${\bf L}^l_t$ as inputs to obtain motion filter ${\bf u}^l_t$;
\STATE Compute response map for each layer, and combine all response maps to obtain confidence map;
\STATE Estimate target location $\bar{bb}_t$ by finding maximum confidence score $s_t$;
\STATE Update motion models using~\eqref{eq::update};
\STATE {\tt // Target re-detection}
\IF{$s_t$ is below $T_1$}
\STATE Generate proposals, and compute their response maps using appearance filter ${\bf w}_{at}$ with ${\bf H}_t$;
\IF{Maximum response score is larger than $T_2$}
\STATE Update $\bar{bb}_t$ as $bb_t$;
\STATE Update appearance models similar to~\eqref{eq::update};
\ENDIF
\ENDIF
\STATE {\tt // Scale estimation}
\STATE Generate a target pyramid, and compute their response maps using scale filter ${\bf w}_{st}$ with ${\bf H}^b_t$;
\IF{Maximum response score is larger than $s_t$}
\STATE Update $\bar{bb}_t$ or $bb_t$ as $bb^*_t$;
\STATE Update scale models similar to~\eqref{eq::update};
\ENDIF
\UNTIL{\emph{End of video sequence}.}
\end{algorithmic}
\end{algorithm}

\section{Tracker Details}
\label{sec::tracking}
Based on the proposed joint learning model, we briefly
present our tracker with four modules, including model updating, target localization, target re-detection and scale handling. Algorithm~\ref{alg::tracking} shows the whole tracking procedure.

\subsection{Tracking Modules}
{\flushleft \bf Model updating}.
To account for appearance changes of target objects, we update the appearance model $\bar{\bf x}$ and the filter model $\bar{\bf u}$ over time. At time $t$, model parameters are updated by:
\begin{equation}\label{eq::update}
\begin{aligned}
&\mathcal{F}(\bar{\bf x})^t=(1-\eta)\mathcal{F}(\bar{\bf x})^{t-1}+\eta\mathcal{F}({\bf x});\\
&\mathcal{F}(\bar{\bf u})^t=(1-\eta)\mathcal{F}(\bar{\bf u})^{t-1}+\eta\mathcal{F}({\bf u}),
\end{aligned}
\end{equation}
where $\eta$ is a learning rate. We update the above models with 3 frames interval to avoiding overfitting.

{\flushleft \bf Target localization}.
Given the learned appearance model $\bar{\bf x}$ and filter model $\bar{\bf u}$, we estimate the target translation by searching for the location of the maximal value of $\bar{\bf y}$ in~\eqref{eq::translation}: 
\begin{equation}\label{eq::translation}
\begin{aligned}
&\bar{\bf y}=\mathcal{F}^{-1}(\mathcal{F}(\bar{\bf u})\odot\sum_d^D\mathcal{F}({\bf x}^d\odot\bar{\bf x}^d)),
\end{aligned}
\end{equation}
where ${\bf x}$ denotes an image patch in the new frame. 

{\flushleft \bf Target re-detection}.
If tracking failures occur, the proposed method is hard to recover targets and would affect
the tracking performance. To handle this problem, we integrate
the scheme of target re-detection into our tracking framework like~\cite{Ma15cvpr,HCF18pami}. Specifically, we set a threshold $T_1$ to judge whether tracking
failures occur or not. If the confidence score is below $T_1$,
we treat the tracker as losing the target and generate a set of
region proposals using the EdgeBox algorithm~\cite{EdgeBox14eccv} across
the whole frame for recovering target objects. Then, another correlation filter learnt over the HOG feature is used to re-detect target objects, and we update this filter with the
learning rate $\eta_2$ when its confidence score is larger a threshold $T_2$.

{\flushleft \bf Scale handling}.
During object tracking, we construct a target pyramid around the estimated translation location for scale estimation~\cite{Ma15cvpr}. Note that  $M\times N$ is the target size in a test frame and let $R$ indicate the number of scales $\mathbb{B}=\{a^{\bar{r}}|\bar{r}=\lfloor-\frac{R-1}{2}\rceil,\lfloor-\frac{R-3}{2}\rceil,...,\lfloor\frac{R-1}{2}\rceil\}$. For each $b\in \mathbb{B}$, we extract an image region of a size $bM\times bN$ centered around the estimated location. Then, we uniformly resize all image regions with the size $M\times N$, and the optimal scale of target can be achieved by evaluating all resized image regions using the
correlation filter learnt over HOG feature for efficiency. The parameter setting of scale
estimation is the same with~\cite{Ma15cvpr}, and we update the scale
filter with the learning rate $\eta_1$.

\subsection{Difference from Previous Work}
It should be noted that our method is significantly different from~\cite{ColorNamesTracker14cvpr,ECO17cvpr} from the following aspects. 1) Danelljan et al.~\cite{ColorNamesTracker14cvpr} compute a matrix via PCA to project high-dimensional features into a lower space, and Danelljan et al.~\cite{ECO17cvpr} formulate a projection matrix into the correlation filter model to project high-dimensional features into low-dimensional ones. While we encode input features on a predefined dictionary to generate new feature representations, and then employ them to optimize the filter in a unified framework. 2) For low-dimensional features (especially for one-dimension, e.g., gray value),~\cite{ColorNamesTracker14cvpr,ECO17cvpr} are not suitable to enhance the discrimination, but our method could handle them and also improve the tracker performance, as demonstrated in the experiments.

Our method is also very different from other feature coding based trackers~\cite{Zhang17tnnls,Liu11cvpr,SCM12cvpr,Zhang13pr,Wang15tip}. These methods usually employ feature coding algorithms to learn a
target-based appearance histogram, and then define the likelihood scores of candidates using similarities with the target template in the Bayesian or particle filtering framework.
Different from these methods, we pursue a robust feature
coding in the correlation filter model to yield a compact,
discriminative and target-oriented feature representation.

\begin{table*}[t]\footnotesize
\caption{PR/SR scores of FOF versus the trackers that only use deep features on the OTB100 dataset, where the best results are in bold fonts. }
\centering
\begin{tabular}{c|c|c c c c c c c c c|c}
\hline
Dataset	& Metric & CFNet & SINT++ & SRDCFdecon & HCF & FCNT & StrutSiam & HDT & DeepSRDCF & MDNet & FOF\\
& & CVPR17 & CVPR18 & CVPR16 & ICCV15 & ICCV15 & ECCV18 & CVPR16 & ICCVW15 & CVPR16  & \\
\hline
OTB50     & PR & 0.807 & 0.839 & 0.870 & 0.891 & 856 & 0.880 & 0.889 & 0.849 & 0.911 & {\bf 0.915}\\
          & SR & 0.611 & 0.624 & 0.653 & 0.605 & 599 & 0.638 & 0.603 & 0.641 & 0.671  & {\bf 0.672} \\
   \hline
	OTB100 & PR & 0.751 & 0.768 & 0.825 & 0.837 & 0.779 & 0.848 & 0.851 & 0.851 & 0.878 & {\bf 0.881} \\
          & SR & 0.580 & 0.574 & 0.627 & 0.562 & 0.551 & 0.564 & 0.621 & 0.635 & {\bf 0.646} & 0.639 \\
\hline
\end{tabular}
\label{tb::otb100}
\end{table*}

\begin{figure*}[t]
  \centering
  \includegraphics[width=0.9\textwidth]{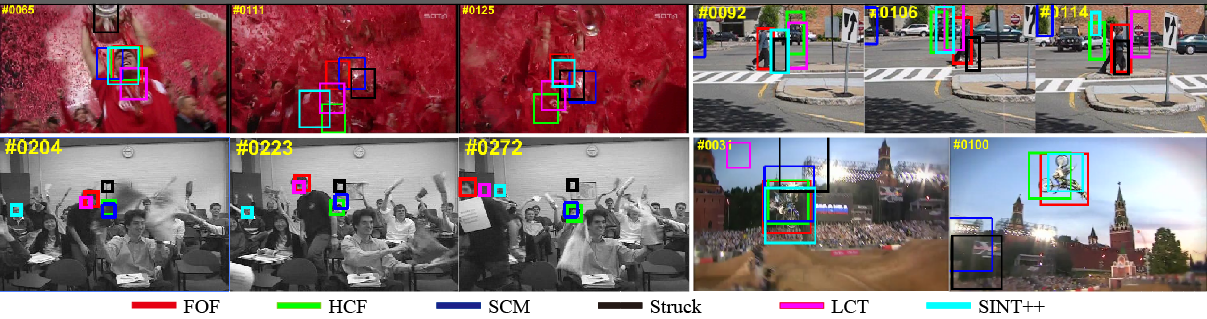} \\
  \caption{Visual examples of our method comparing five trackers on four video sequences. }\label{fig::visual_results}
\end{figure*}

\section{Performance Evaluation}
\label{sec::evaluation}
To validate
the effectiveness of our framework, i.e., Filter Optimization
driven Feature coding (FOF), we evaluate it on three benchmarks, i.e., the OTB50 dataset~\cite{RGBbenchmark13cvpr}, the OTB100 dataset~\cite{RGBbenchmark15pami} and the VOT2016
dataset~\cite{vot16challenge}. At last, we analyze the proposed model.

\subsection{Evaluation Setting}

{\flushleft \bf Implementation details}.
We adopt VGGNet-19 trained
on the ImageNet dataset for feature extraction, and use the
outputs of the \emph{conv3-4}, \emph{conv4-4} and \emph{conv5-4} convolutional
layer as our features. The proposed algorithm is employed
on these features and their response maps are combined together with the weights 0.25, 0.5 and 1, respectively. We
keep the learning rates $\eta$, $\eta_1$ and $\eta_2$ the same as 0.01 for simplicity, and
set the thresholds $T_1$ and $T_2$ as 0.25 and 0.38, respectively.
For generating proposals in EdgeBox, we set the step size
to 0.85 and the NMS (non-maximum suppression) threshold to 0.8. In the proposed model~\eqref{eq::objective}, we empirically set
$\lambda=0.5$ and $\gamma=0.8$.

{\flushleft \bf Evaluation metrics}.
On both OTB50 and OTB100 datasets~\cite{RGBbenchmark13cvpr,RGBbenchmark15pami}, we use
precision rate (PR) and success rate (SR) for quantitative
performance. PR is the percentage of frames whose output
location is within a threshold distance of ground truth, and
SR is the percentage of the frames whose overlap ratio between the output bounding box and the ground truth bounding box is larger than a threshold. We set the threshold to be
20 pixels to obtain the representative PR, and employ the area under the curves of success rate as the representative
SR for quantitative performance.
On the VOT2016 dataset~\cite{vot16challenge}, we adopt 3 primary measures (i.e., Accuracy (A), robustness (R) and expected average overlap (EAO)) to assess a tracker.
A is the average overlap between the predicted and ground
truth bounding boxes during successful tracking periods,
and R the robustness measures how many times the tracker
loses the target (fails) during tracking. EAO is an estimator
of the average overlap a tracker is expected to attain on a
large collection of short-term sequences with the same visual properties as the given dataset~\cite{vot16challenge}.

\subsection{Evaluation on the OTB50 Dataset}
On the OTB50 dataset, we evaluate our approach with
comparison to nine state-of-the-art trackers that only use
deep learning features, including MDNet~\cite{MDNet15cvpr}, DeepSRDCF~\cite{DeepSRDCF15iccvw}, HDT~\cite{HDT16cvpr}, StrutSiam~\cite{StrutSiam18eccv}, FCNT~\cite{Wang15iccv},
HCF~\cite{Ma15iccv}, SRDCFdecon~\cite{SRDCFdecon16cvpr}, SINT++~\cite{SINT++18cvpr}, and CFNet~\cite{CFNet17cvpr}.
Table~\ref{tb::otb100} shows the results, which suggest that our FOF
generally perform well against the state-of-the-art trackers
on the OTB50 dataset. In particular, our method improves
the baseline HCF by a large margin (2.4\%/6.7\% performance gains in PR/SR), and outperforms the state-of-the-art MDNet, which is not
offline trained on auxiliary sequences for fair comparison.
The overall promising performance of our method can be
explained by the fact that the proposed joint learning algorithm strengthens the discriminative power of the filter by
suppressing feature redundancy and noise.

We also present four visual examples of our method comparing five trackers in Fig.~\ref{fig::visual_results}, including HCF~\cite{Ma15iccv}, SCM~\cite{SCM12cvpr}, Struck~\cite{Stuck11iccv}, LCT~\cite{Ma15cvpr} and SINT++~\cite{SINT++18cvpr}, which qualitatively justify the effectiveness of our FOF tracker in handling the challenges of
motion blur, low resolution, partial occlusion, appearance variation, target rotation and background clutter. 

\subsection{Evaluation on the OTB100 Dataset}
On the OTB100 dataset, we also evaluate our approach with
the above nine state-of-the-art trackers, as shown in Table~\ref{tb::otb100}.
The results on the OTB100 dataset again demonstrate the similar observations with the OTB50 dataset. Specifically, our FOF outperforms
the baseline HCF by 4.4\%/7.7\% in PR/SR, bigger performance gains than on the OTB50 dataset. Comparing with MDNet, we achieve superior performance in PR, but slightly worse in SR. It is because our tracker uses a simple strategy that samples a sparsely set of scaled regions from an image pyramid and then evaluates them using a HOG-based correlation filter for scale estimation. While the MDNet method adopts the bounding box regression model trained with deep features to improve the
target localization accuracy. Overall, the favorable results against the state-of-the-art methods on the OTB100 dataset further demonstrate the effectiveness of the proposed approach.

\begin{table}[t]\footnotesize
\caption{Accuracy, Robustness and EAO on the VOT2016 dataset, where the best results are in bold fonts. }
\centering
\begin{tabular}{c|c c c c |c}
\hline
	 & HCF & SiamFC & SRDCF & MDNet & FOF\\
   \hline
	A & 0.445 & 0.527 & 0.532 & {\bf 0.541} & 0.531 \\
   R & 0.664 & 0.630 & 0.657 & 0.714 & {\bf 0.760} \\
   EAO & 0.220 & 0.235 & 0.247 & 0.257 & {\bf 0.307} \\
\hline
\end{tabular}
\label{tb::vot}
\end{table}

\subsection{Evaluation on the VOT2016 Dataset}
Finally, we report the evaluation results of FOF against MDNet~\cite{MDNet15cvpr}, SRDCF~\cite{danelljan2015learning}, SiamFC~\cite{SiamFC16eccv} and HCF~\cite{Ma15iccv} on the VOT2016 dataset~\cite{vot16challenge}, as shown in Table~\ref{tb::vot}. From the results we can see that, the performance of our FOF is clearly better than MDNet, SRDCF, SiamFC and HCF in terms of most metrics, further demonstrating the effectiveness of the proposed tracker. The overlap ratios of our method is lower than MDNet, and we have explain the reason in the analysis on the OTB100 dataset.  The VOT2016
report suggests that trackers whose EAO value exceeds 0.251 belong to the state-of-the-art, and so our FOF is the state-of-the-art.

\begin{table}[t]\footnotesize
\caption{Analysis of parameter sensitivity on the OTB100 dataset. }
\centering
\begin{tabular}{c| c c c| c c c  }
\hline
	 & \multicolumn{3}{c|}{$\lambda$} & \multicolumn{3}{c}{$\gamma$}  \\
\hline
 &  0.3 & 0.5 & 0.7 & 5 & 10 & 15 \\
   \hline
	PR & 0.871 & 0.881 & 0.879 & 0.878 & 0.881 & 0.879   \\
   SR & 0.635 & 0.639 & 0.638 & 0.637 & 0.639 & 0.637  \\
\hline
\end{tabular}
\label{tb::parameter_sensitivity}
\end{table}

\begin{figure}[t]
  \centering
  \includegraphics[width=\columnwidth]{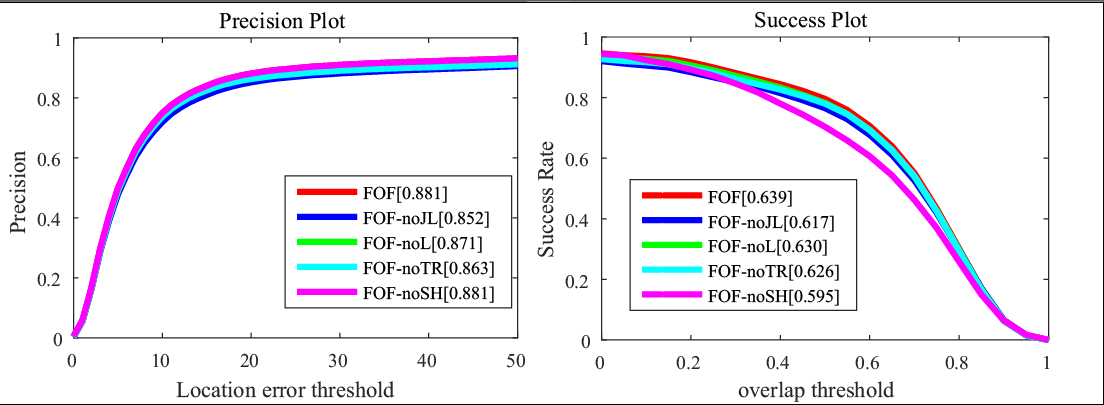} \\
  \caption{Ablation study on the OTB100 dataset.}\label{fig::component}
\end{figure}

\subsection{In-depth Analysis of Our Approach}
Our implementation in MATLAB runs on a PC with an i7-4.2 GHz
CPU with 32GB memory. 
To clarify the proposed approach, we analyze it in detail from the following four aspects.

{\flushleft \bf Sensitivity analysis of model parameters}.
The scalar parameters $\lambda$ and $\gamma$ in~\eqref{eq::objective} are to avoid overfitting of ${\bf u}$ and make a balance between the Laplacian term and other terms. We set $\lambda$ to 0.3, 0.5 and 0.7, and $\gamma$ to 5, 10 and 15 to evaluate the tracking performance on the OTB100 dataset. Table~\ref{tb::parameter_sensitivity} shows the results, and the tracking performance is not disturbed much when slightly adjusting $\lambda$ and $\gamma$.

{\flushleft \bf Ablation study}.
We conduct experiments to justify the effectiveness of the used components in our approach. They are: 1) noJL, that removes the scheme of joint learning in~\eqref{eq::objective}, i.e., first performing feature coding then employing the coding features to train the correlation filter; 2) noL, that removes the Laplacian constraint on feature coding in~\eqref{eq::objective}; 3) noTR, that removes the scheme of target re-detection in the proposed tracking method; and 4) noSH, that removes the strategy of scale handling in our tracking approach. The results are presented in Fig.~\ref{fig::component}. We can see that the proposed joint scheme plays a very critical role in improving the tracking performance by observing the large improvement (2.9\%/2.2\% in PR/SR) of FOF over FOF-noJR. The Laplacian constraint is also helpful in enhancing the coding quality and stability. In addition, the target re-detection scheme and the scale handling strategy improve the tracking performance considerably. Overall, the results justify the effectiveness of different components introduced in our tracking framework.

\begin{table}[t]\footnotesize
\caption{Analysis of parameter sensitivity on the OTB100 dataset. }
\centering
\begin{tabular}{c| c c c| c c }
\hline
	 & \multicolumn{3}{c|}{{\bf Size}} & \multicolumn{2}{c}{{\bf Algorithm}}  \\
\hline
 &  3 & 10 & 20 & $k$-means & $k$-means\\
   \hline
	PR & 0.878 & 0.881 & 0.872 & 0.853 & 0.863  \\
   SR & 0.636 & 0.639 & 0.633 & 0.624 & 0.628 \\
\hline
\end{tabular}
\label{tb::codebook}
\end{table}

{\flushleft \bf Impacts of codebook}.
We also study the impact of different sizes of codebook, and the results are shown in Table~\ref{tb::codebook}. The results show that the performance raises or drops a little when increasing or decreasing the codebook size, and thus we set it to 10. It can be explained by the fact that the smaller size of codebook decreases the diversity of the learnt features, and the larger size will increase the instability of feature codes, which could be solved by introducing more prior constraints on the learnt features, and we will study it in the future. In addition, we replace the dictionary learning algorithm with the $k$-means algorithm to construct the codebook, and find that the tracker performance is affected much and the results of two runs are different due to the random initialization of $k$-means. It again demonstrates the effectiveness of the dictionary learning scheme adopted in our method.

{\flushleft \bf Efficiency analysis}.
Finally, we present the runtime of our FOF using a single CPU against the baseline HCF~\cite{Ma15iccv} and state-of-the-art MDNet~\cite{MDNet15cvpr} with their tracking performance on the OTB100 dataset in Table~\ref{tb::efficiency}. Overall, the results demonstrate that our framework
clearly outperforms baseline HCF (4.1\%/7.7\% performance gains in PR/SR) with a modest impact
on the frame rate (1.86 FPS versus 2.36 FPS), and performs comparably against the state-of-the-art MDNet but much faster than it (1.86 FPS versus 0.27 FPS). We also report the time costs of major parts of our approach, and the time costs of feature extraction, target re-detection, scale handling and the optimization occupy 67\%, 15\%, 10\% and 8\%, respectively. One can see that the feature extraction costs most time, and the proposed optimization algorithm only consumes a little (8\%, about 40ms per frame on a single CPU).

\begin{table}[t]\footnotesize
\caption{Performance and runtime of our FOF against the baseline and state-of-the-art methods on a single CPU. }
\centering
\begin{tabular}{c| c c c }
\hline
	 & HCF & MDNet & FOF  \\
   \hline
	PR & 0.837 & 0.878 & 0.881  \\
   SR & 0.562 & 0.646 & 0.639 \\
\hline 
  FPS & 2.36 & 0.27 & 1.86 \\
\hline
\end{tabular}
\label{tb::efficiency}
\end{table}

\section{Conclusion}
\label{sec::conclusion}
In this paper, we have proposed a joint learning algorithm to enhance the discriminative capacity of the correlation filter, which outperforms the baseline trackers with a clear margin and also achieves favorable performance against the state-of-the-art trackers on three tracking datasets. The proposed algorithm bridges the feature coding and correlation filter learning, and provides a potential research direction to the visual tracking task. In future work, we will integrate other feature coding algorithms and spatio-temporal cues into our framework to improve the robustness of representation learning and thus enhance the discriminative capacity of the correlation filter. 


{\small
\bibliographystyle{ieee}
\bibliography{mybibfiles}
}

\end{document}